\pdfoutput=1
\documentclass[11pt]{article}

\usepackage[]{acl}
\usepackage{pifont}
\usepackage{amsmath}
\usepackage{graphicx}
\usepackage{multirow}
\usepackage{array}
\usepackage{times}
\usepackage{latexsym}

\usepackage[T1]{fontenc}
\usepackage[utf8]{inputenc}

\usepackage{microtype}

\usepackage{inconsolata}

\usepackage{graphicx}

\title{Learning by Analogy: Enhancing Few-Shot Prompting for Math Word Problem Solving with Computational Graph-Based Retrieval}

\author{
 \textbf{Xiaocong Yang\textsuperscript{1}},
 \textbf{Jiacheng Lin\textsuperscript{1}},
 \textbf{Ziqi Wang\textsuperscript{1}},
 \textbf{ChengXiang Zhai\textsuperscript{1}}
 \\
 \textsuperscript{1}University of Illinois Urbana-Champaign
 \\
\texttt{\{xy51,jl254,ziqiw9,czhai\}@illinois.edu}
}

\begin{document}
\maketitle
\begin{abstract}
Large language models (LLMs) are known to struggle with complicated reasoning tasks such as math word problems (MWPs). In this paper, we present how analogy from similarly structured questions can improve LLMs’ problem-solving capabilities for MWPs. Specifically, we rely on the retrieval of problems with similar \textit{computational graphs} to the given question to serve as exemplars in the prompt, providing the \textit{correct reasoning path} for the generation model to refer to. Empirical results across six math word problem datasets demonstrate the effectiveness of our proposed method, which achieves a significant improvement of up to 6.7 percent on average in absolute value, compared to baseline methods. These results highlight our method's potential in addressing the reasoning challenges in current LLMs.

% With our approach, the exact match (EM) score on six math word problem tasks increases by up to 6.7 percent in absolute value on average compared to the BGE baseline. Our work provides a promising solution to the lack of reasoning ability in current LLMs.
\end{abstract} 

\section{Introduction}

Large Language Models (LLMs) have demonstrated remarkable success across a wide range of tasks \cite{achiam2023gpt, dubey2024llama3herdmodels, jiang2023mistral, labrak2024biomistral, lin2024panacea}. However, solving math word problems (MWPs) remains a significant challenge for LLMs \cite{ahn2024large, srivatsa2024makes}. Unlike tasks that primarily rely on linguistic or general knowledge, MWPs demand a nuanced integration of language comprehension and mathematical reasoning, posing unique difficulties for LLMs. Overcoming this challenge is critical, as proficiency in solving MWPs could expand the applications of LLMs to education, automated tutoring, and complex reasoning tasks.

Human problem-solving for MWPs offers an insightful source of inspiration. People often solve new problems by analogy, leveraging prior examples to adapt solutions to novel scenarios. Inspired by this analogy-driven learning process, recent research has employed few-shot prompting techniques to enhance MWP performance in LLMs \cite{jiang2023mistral, melz2023enhancing, henkel2024retrieval}. Most existing approaches for selecting few-shot examples rely either on random selection \cite{jiang2023mistral, dubey2024llama3herdmodels} or retrieval based solely on semantic similarity \cite{huang2023boosting, melz2023enhancing, henkel2024retrieval}. Although providing examples can improve LLM performance, these methods often fail to ensure that the selected examples align with the mathematical structure of the target problem. Specifically, randomly selected examples lack relevance to the target problem, while semantic retrieval tends to prioritize superficial linguistic similarity over deep structural alignment. This mismatch between the provided examples and the target problem ultimately constrains the effectiveness of LLMs in solving MWPs.

To address this limitation, we propose a novel computational graph-based retrieval method for selecting examples that align more closely with the underlying structure of the target math word problem. Our approach identifies examples with computational graphs that are structurally similar to the target problem and incorporates these examples into few-shot prompting, providing the LLM with more relevant problem-solving guidance. Specifically, we design a lightweight retriever model trained using contrastive learning to identify structurally analogous examples. Examples with similar graphs are treated as positive pairs, while those with dissimilar graphs are treated as negative pairs. Once trained, the retriever can be seamlessly integrated into the LLM inference workflow without requiring updates to the LLM’s parameters, making our approach modular and easily adaptable. We evaluate our method on six math word problem datasets, demonstrating that our computational graph-based retrieval approach achieves significant performance improvements over both semantic-based retrieval and random selection baselines. Furthermore, we conduct case studies and detailed analyses to highlight the effectiveness of our method.

\begin{figure}[t]
    \centering
    \includegraphics[width=0.48\textwidth]{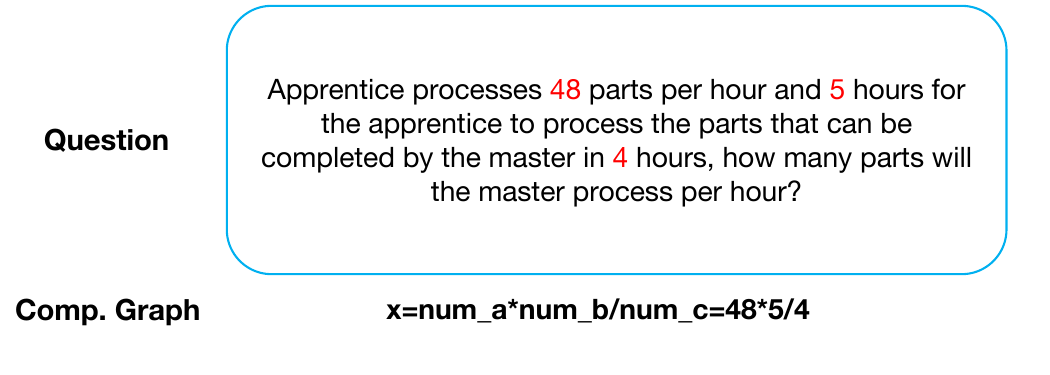} 
    \caption{An example of a math word problem with its computational graph.}
    \label{fig:comp_graph}
    \vspace{-1em}
\end{figure}

Our contributions are summarized as follows:
\begin{itemize}
    \item \textbf{Proposing Computational Graph-Based Retrieval for Few-Shot Prompting.} We introduce a computational graph-based retrieval method specifically tailored for math word problem-solving. This approach selects examples with structural similarity to the target problem, enhancing few-shot prompting by providing LLMs with examples that align with the underlying mathematical structure of the problem.

    \item \textbf{Training a Structural Similarity Retriever.} We develop a retriever model trained with contrastive learning to identify structural similarity in math word problems. This lightweight and modular retriever integrates seamlessly into the LLM inference workflow without requiring parameter updates to the LLM itself.
    
    \item \textbf{Conducting Extensive Evaluation and Analysis.} We conduct comprehensive experiments on six math word problem datasets, demonstrating that our approach significantly outperforms both semantic-based and random selection baselines, with average exact matching (EM) score improvements of up to 6.7\% and 19.5\% respectively. Additionally, we present in-depth case studies and analyses to validate the effectiveness of our method in capturing structural nuances essential for MWP-solving. We also provide an automated approach to construct the training data without any human labors.
\end{itemize}
\section{Methodology}

\begin{figure*}[htbp]
    \centering
    \includegraphics[width=1.0\textwidth]{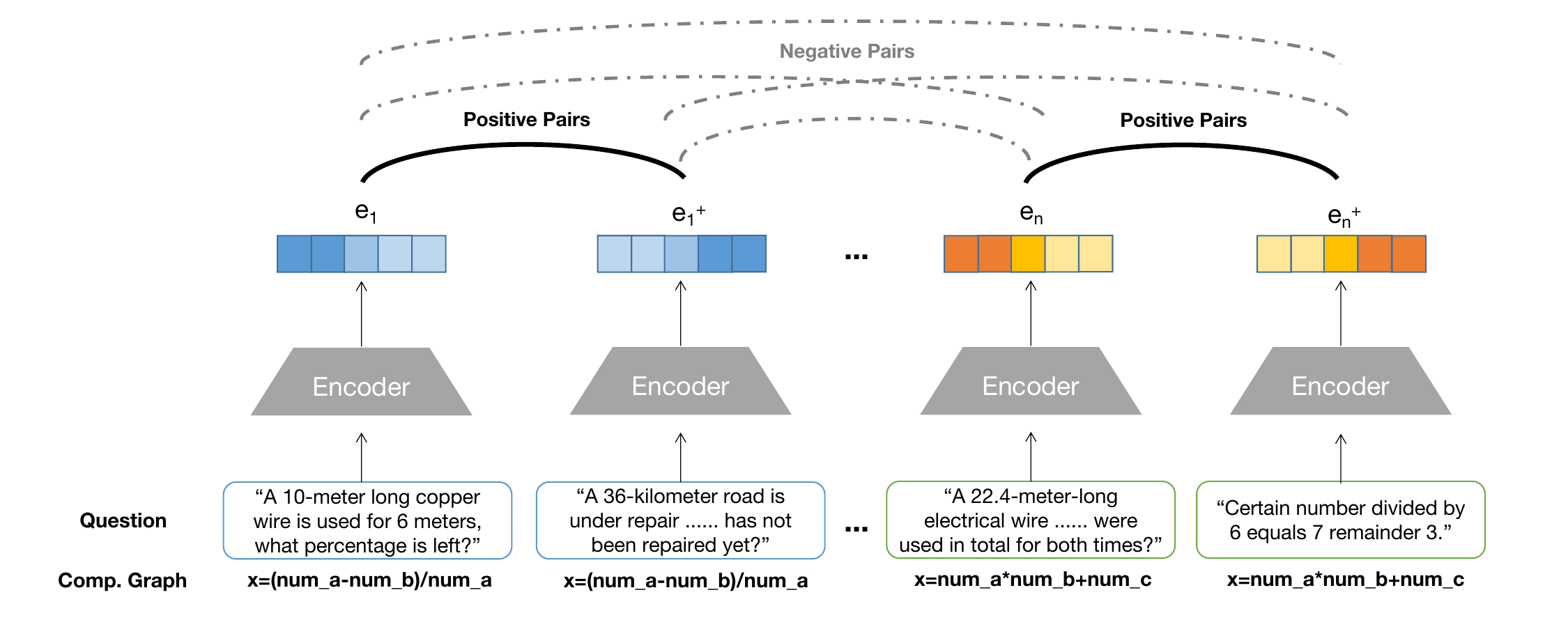} 
    \caption{\textbf{Flowchart of Retriver Training.} This figure illustrates the process of training a retriever model (encoder) with contrastive learning to identify structurally similar math word problems. Each question is encoded into an embedding based on its text. Positive pairs are formed by pairing examples with matching computational graph structures, while in-batch negatives serve as contrasting examples with different structures. }
    \label{fig:main_graph}
\end{figure*}

\subsection{Overview of the Proposed Framework}

When solving a new reasoning problem, humans often draw upon known problems with similar reasoning paths and address them by analogy. In the context of math word problems, the reasoning path corresponds to its computational graph, as illustrated in Figure \ref{fig:comp_graph}. Large language models (LLMs) are observed to fail to conduct genuine logical reasoning \cite{mirzadeh2024gsmsymbolicunderstandinglimitationsmathematical} and exhibit strong token biases \cite{li-etal-2024-deceptive} when addressing reasoning tasks. Therefore, providing LLMs with the \textit{correct reasoning path} from analogous problems can guide them to mimic the problem-solving process. This paper aims to develop a math word problem-solving system comprising a retriever and a generator. The retriever identifies problems and solutions with computational graphs similar to the query problem from a corpus, while the generator leverages these retrieved exemplars through in-context learning to enhance problem-solving performance.

\subsection{Retriever Model Training}
Figure \ref{fig:main_graph} shows the training process of the retriever. Given a batch of questions $\{q_i\}_{i=1}^n$ and their corresponding computational graphs $\{G_i\}_{i=1}^n$, we search in the training dataset for positive examples $\{q_i^+\}_{i=1}^n$ where their computational graphs are the same as those of the query questions: $G_i^+ = G_i$, $i=1,2,...n$ where $n$ is the batch size.\footnote{We discard the examples if there's no positive samples matched in the training dataset.} Then we forward the $\{q_i,q_i^+\}_{i=1}^n$ with the retriever (an encoder model) $f_{\theta r}$ to get the embeddings $\{f_{\theta r} (q_i),f_{\theta r} (q_i^+)\}_{i=1}^n$. By applying infoNCE loss \cite{oord2018representation} with the in-batch negative strategy, the training loss objective $L$ of the retriever becomes:
\begin{align}
    L &= \frac{1}{n}\sum_{i=1}^{n}  -\log ( e^{\text{sim}(f_{\theta r} (q_i), f_{\theta r} (q_i^+)) / \tau} / \nonumber  \\
    & ( \sum_{j=1, j \neq i}^{n} e^{\text{sim}(f_{\theta r} (q_i), f_{\theta r} (q_j)) / \tau} + \nonumber \\
    & \sum_{j=1}^{n} e^{\text{sim}(f_{\theta r} (q_i), f_{\theta r} (q_j^+)) / \tau}))
\end{align}
where ${\rm sim}$ indicates a similarity function and $\tau$ is the temperature. Note that we do not need to train the generator.

% \begin{align}
%     l_i 
%     = & - \log \left( {e^{\text{sim}(f_{\theta r} (q_i), f_{\theta r} (q_i^+)) / \tau}} / \left( \right. \notag \\
%     & {\sum_{j \neq i}^{n} e^{\text{sim}(f_{\theta r} (q_i), f_{\theta r} (q_j)) / \tau}}\left. + \sum_{j=1}^{n} e^{\text{sim}(f_{\theta r} (q_i), f_{\theta r} (q_j^+)) / \tau}\right) \right)
% \end{align}

% \begin{equation}
%     L = \frac{1}{n}\sum_{i=1}^{n}l_i
% \end{equation}

\subsection{Inference}
 Given a trained retriever $f_{\theta r^*} $, a question-solution pair corpus $C$ and a given question $q$, the retriever select the top-k similar question-solution pairs $\{q_i,a_i\}_{i=1}^k$ based on the similarity score:
\begin{equation}
    \{q_i,a_i\}_{i=1}^k = topk(\text{sim}(f_{\theta r^*} (q), f_{\theta r^*} (q_j)))
\end{equation}
where $q_j \in C$. Then we concatenate the retrieved question-answer pairs and the given question as the prompt to the generator $f_{\theta g}$ to get the output answer $a$:
\begin{equation}
    a = f_{\theta g} (concat(q_1,a_1,...,q_k,a_k,q))
\end{equation}
where $concat$ denotes the concatenation operation.

\section{Experiment}

\subsection{Setup}

\paragraph{Implementation Details.} In our experiments, we use BGE-large-en-v1.5 \cite{bge_embedding} as retriever and LLaMA-3 model series \cite{dubey2024llama3herdmodels} as generator for English datasets (except for 0.5B size experiments, where we use Qwen2.5-0.5B-Instruct as the generator since no similar sized LLaMA-3 model is available), and BGE-large-zh-v1.5 as retriever and Qwen2.5 model series \cite{qwen2.5} as generator for Chinese datasets, with bfloat16 precision for all models. We add an extra pooler (a two-layer MLP module) to the retriever, following the practice in \cite{chen2020simpleframeworkcontrastivelearning}. We train the retriever on 25\% randomly selected data from Math23k training set\footnote{Math23k dataset is provided in Chinese, and we use LLaMA-3.1-70B-Instruct to translate it into English for the training of English model.} \cite{wang-etal-2017-deep} where the computational graphs are provided, using AdamW \cite{loshchilov2019decoupledweightdecayregularization} with a learning rate of 3e-5 for 5 epochs, a temperature $\tau$ of 0.05, and cosine similarity as the similarity function. We set the batch size equal to 16 for the training process.

\paragraph{Datasets.} We evaluate our retrieval-generation system on the following six math word problem datasets: Math23k \cite{wang-etal-2017-deep}, ape210k \cite{zhao2020ape210klargescaletemplaterichdataset}, gsm8k \cite{cobbe2021trainingverifierssolvemath}, math\_qa \cite{amini-etal-2019-mathqa}, Calc-ape210k \cite{kadlcik-etal-2023-soft} and aqua\_rat \cite{ling2017program}, as shown in Table \ref{table:datsets}. For all datasets, we use the corresponding training set as the retrieval corpus and evaluate on the test set, and use $k=8$ for top-k example retrieval. 
\begin{table*}[ht]
\centering
\renewcommand{\arraystretch}{1.2} % 设置行距为1.2倍
\begin{tabular}{lccccc}
\hline
& \textbf{\# Samples (train/val/test)} & \textbf{Language} & \textbf{Solution type} & \textbf{Comp. Graph}  & \textbf{Options} \\
\hline
\textbf{Math23k} & 21.2k/1k/1k &ZH& Equation & \ding{51}  & \ding{55}\\
\textbf{ape210k}& 200.5k/5k/5k &ZH& Equation & \ding{55}  & \ding{55}\\
\textbf{gsm8k}& 7.5k/-/1.3k &EN& Text & \ding{55}  & \ding{55}\\
\textbf{math\_qa}& 29.8k/4.5k/3.0k &EN& Text & \ding{55} & \ding{51}\\
\textbf{Calc-ape210k}& 195k/1.8k/1.8k &EN& Equation & \ding{55} & \ding{55}\\
\textbf{aqua\_rat}& 97.5k/254/254 &EN& Text & \ding{55} & \ding{51}\\
\hline
\end{tabular}
\caption{Details of datasets evaluated.``ZH'' and ``EN'' refers to Chinese and English. An example of equation solution is ``\textit{x=(5*1000)-2000}'' where $x$ is the final answer, and an example of text solution is ``\textit{Natalia sold 48/2 = <<48/2=24>>24 clips in May. Natalia sold 48+24 = <<48+24=72>>72 clips altogether in April and May. \#\#\#\# 72.}''Options refer to if the candidate answers are provided in the question.}
\label{table:datsets}
\end{table*}

\paragraph{Metrics.} We report exact match (EM) accuracy for all datasets. During inference, we require the generator to generate answers following the same format of the given exemplars to facilitate the parsing of the solution to obtain the final answer, and consider the generated solution correct if the parsed final answer matches the golden answer. We use string matching for the datasets where the solutions are provided in text format, and use float number matching if the solutions are provided in equation format.

\begin{table*}[ht]
\centering
\renewcommand{\arraystretch}{1.2}
\resizebox{\linewidth}{!}{
\begin{tabular}{lccccccc}
\hline
\textbf{} & \textbf{Math23k} & \textbf{ape210k} & \textbf{gsm8k} & \textbf{math\_qa} & \textbf{Calc-ape210k} & \textbf{aqua\_rat} & \textbf{Avg.} \\
\hline
\textbf{Random\textsubscript{Qwen-0.5B}} & 28.9 & 19.2 & 17.1 & 16.5 & 12.0 & 18.1 & 18.6 \\
\textbf{BGE\textsubscript{Qwen-0.5B}}  & 43.1 & 39.7 & 21.2 & \textbf{27.3} & 17.6 & 16.9 & 27.6\\
\textbf{Ours\textsubscript{Qwen-0.5B}} & \textbf{57.6} & \textbf{49.2} & \textbf{22.7} & 26.6 & \textbf{30.5}& \textbf{18.9} & \textbf{34.3} \\
\hline
\textbf{Random\textsubscript{LLaMA-1B/Qwen-1.5B}} & 50.3 & 32.7 & 38.6 & 17.2 & 22.8 & 14.2 & 27.6 \\
\textbf{BGE\textsubscript{LLaMA-1B/Qwen-1.5B}} & 58.7 & 50.4 & 38.7 & 45.9 & 20.4 & 29.9 & 40.7 \\
\textbf{Ours\textsubscript{LLaMA-1B/Qwen-1.5B}} & \textbf{66.6} & \textbf{59.2} & \textbf{40.7} & \textbf{47.3} & \textbf{31.3} & \textbf{37.4} & \textbf{47.1} \\
\hline
\textbf{Random\textsubscript{LLaMA-3B/Qwen-3B}} & 68.0 & 44.3 & 71.4 & 52.9 & 32.6 & 46.9 & 52.7 \\
\textbf{BGE\textsubscript{LLaMA-3B/Qwen-3B}} & 73.1 & 54.6 & 71.5 & \textbf{64.9} & 31.5 & 50.0 & 57.6  \\
\textbf{Ours\textsubscript{LLaMA-3B/Qwen-3B}}  & \textbf{78.3} & \textbf{59.9} & \textbf{71.9} & 64.3 & \textbf{39.8} & \textbf{50.6} & \textbf{60.8}\\ 
\hline
\textbf{Random\textsubscript{LLaMA-8B/Qwen-7B}} & 83.9 & 62.8 & 80.1 & 51.3 & 30.6 & 49.6 & 59.7 \\
\textbf{BGE\textsubscript{LLaMA-8B/Qwen-7B}} & 87.6 & 73.8 & \textbf{80.4} & 66.4 & 39.5  &  49.6& 66.2 \\
\textbf{Ours\textsubscript{LLaMA-8B/Qwen-7B}} & \textbf{90.4} & \textbf{76.7} & 79.2 & \textbf{66.8} & \textbf{46.5} &\textbf{53.1} & \textbf{68.8} \\
\hline
\textbf{Random\textsubscript{LLaMA-70B/Qwen-72B}} & 84.7 & 68.9 & 84.7 & 60.6 & 39.3 & 59.8 & 66.3 \\
\textbf{BGE\textsubscript{LLaMA-70B/Qwen-72B}} & 90.9 & 79.5 & 86.0 & \textbf{68.5} & 47.9 & \textbf{64.2} & 72.8\\
\textbf{Ours\textsubscript{LLaMA-70B/Qwen-72B}} & \textbf{92.4} & \textbf{80.9} & \textbf{87.3} & 68.0 & \textbf{53.5} & \textbf{64.2} & \textbf{74.4} \\
\hline
\end{tabular}
}
\caption{Main results of our system. We report exact match (EM) for all tasks. Our approach outperforms the baselines on most tasks except for math\_qa, which is because the semantic similarity and computational graph similarity are overlapped in this dataset. While our method is effective for generators of all sizes, the performance gain is larger for smaller models.}
\label{table:main_result}
\end{table*}

\subsection{Main Results}
Table \ref{table:main_result} presents a detailed summary of our experimental results, highlighting the superiority of our method across various datasets and model sizes. Specifically, for the Chinese datasets Math23k and ape210k, our approach consistently and significantly outperforms both the random and BGE baselines. Similarly, strong performance gains are observed across four English datasets, further demonstrating the effectiveness of our method. The only exception is the math\_qa dataset, where our method performs comparably to the BGE baseline. This anomaly arises because, in math\_qa, the semantic similarity often coincides with computational graph similarity. Many example pairs in this dataset differ only in the numerical values while maintaining identical semantic structures and computational graphs (e.g., \textit{“The banker’s gain of a certain sum due 3 years hence at 10\% per annum is Rs. 36. What is the present worth?”} and \textit{“The banker’s gain of a certain sum due 2 years hence at 10\% per annum is Rs. 24. What is the present worth?”}). Since these pairs exhibit similar semantics and identical computational graphs at the same time, the BGE model can effectively retrieve them by focusing solely on semantic similarity, leaving little room for improvement through retriever training. Furthermore, our method demonstrates larger performance gains when the generator model is smaller in size. This could be attributed to the enhanced reasoning capabilities of larger LLMs, which allow them to solve problems more independently, reducing their reliance on retrieving similar examples.

\begin{table}[ht]
\centering
\renewcommand{\arraystretch}{1.2} % 设置行距为1.2倍
\begin{tabular}{lccc}
\hline
& \textbf{Ours} & \textbf{BGE} & \textbf{Upper Bound} \\
\hline
\textbf{Math23k} & 66.6 & 58.7 & \textbf{68.2} \\
\hline
\end{tabular}
\caption{Comparison of our methods with the upper bound with Qwen2.5 1.5B model. Our approach results in a large performance gain compared to the original BGE model and a score close to the upper bound, suggesting the effectiveness of our training process.}
\label{table:upper_bound}
\end{table}

\subsection{Analysis}

\subsubsection{The Performance Upper Bound}
In this work, we hypothesize that problems with similar computational graphs can facilitate answering the given question. Under this assumption, the upper bound of our method’s performance is achieved by using computational graphs directly for retrieval. Since computational graphs are available only on Math23k dataset, we focus on this dataset to compare the upper bound performance with performance of our trained retriever, thereby evaluating the quality of the retriever training process. To measure similarity between computational graphs for retrieval, we utilize the normalized Levenshtein Distance, which quantifies the string-based similarity of computational graph representations. Table \ref{table:upper_bound} compares the performance of our method against the hypothesized upper bound. The results indicate that, compared to the original BGE model, our trained retriever achieves performance significantly closer to the upper bound. This highlights the effectiveness of our training approach in improving retrieval quality.

\subsubsection{Case Study on Retrieved Data}

\begin{figure*}[htbp]
    \centering
    \includegraphics[width=0.9\textwidth]{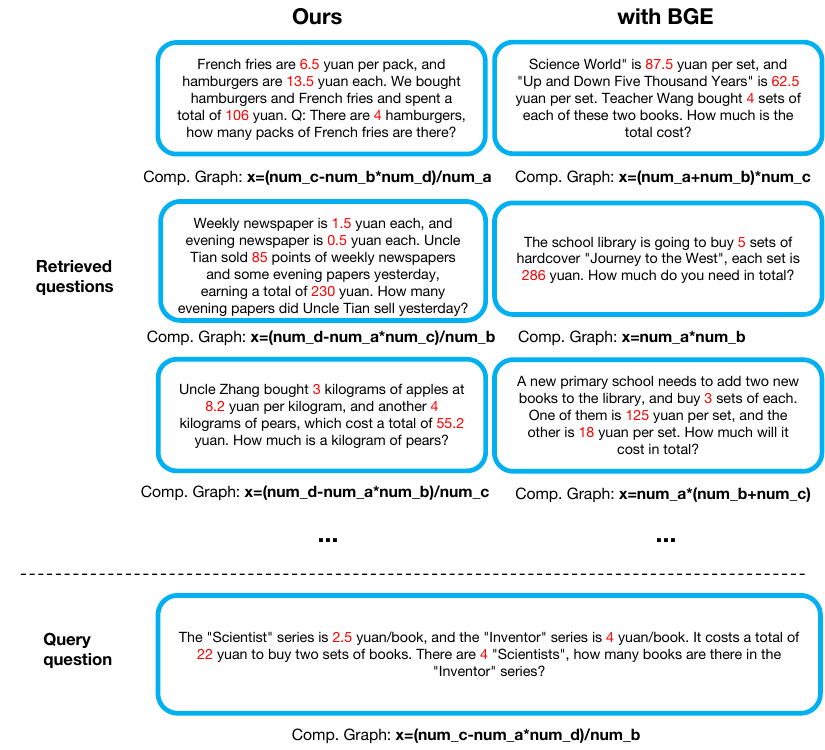} 
    \caption{Case study on the retrieved data with our model and BGE respectively. The retrieved data using trained retriever have similar computational graphs with the query question, while the computational graphs are different for retrieved data using BGE model. }
    \label{fig:case_study}
\end{figure*}

\begin{figure*}[htbp]
    \centering
    \begin{minipage}{0.48\textwidth}
        \centering
        \includegraphics[width=\linewidth]{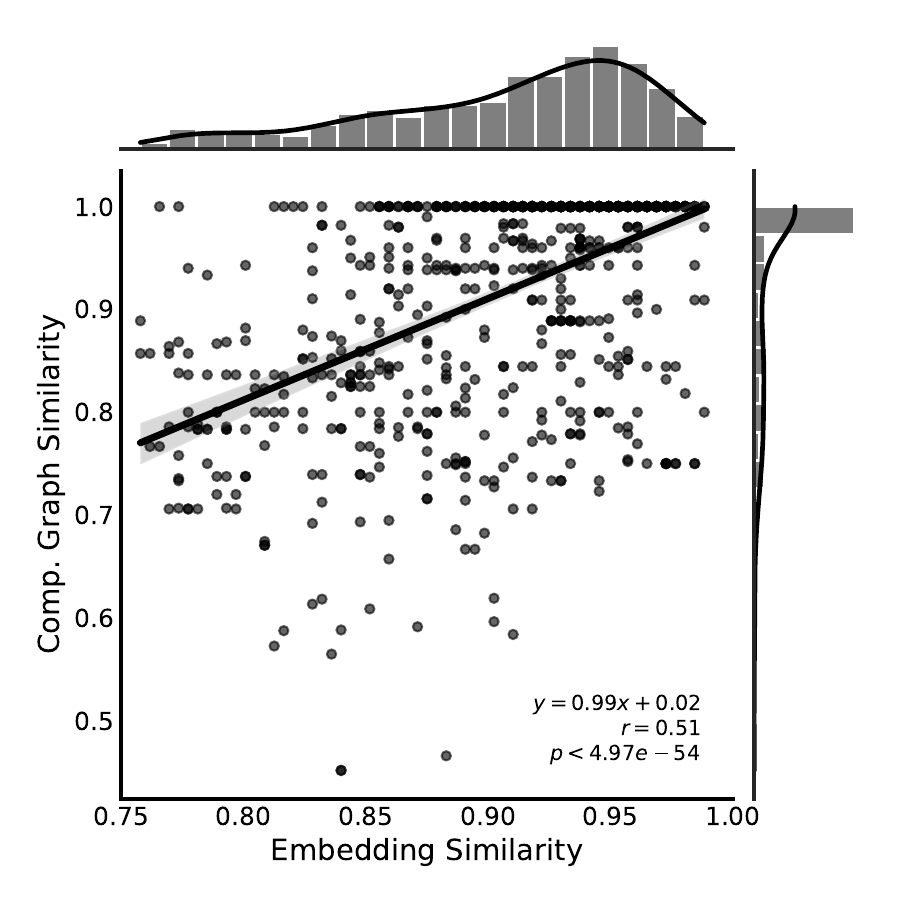}
    \end{minipage}
    \hfill
    \begin{minipage}{0.48\textwidth}
        \centering
        \includegraphics[width=\linewidth]{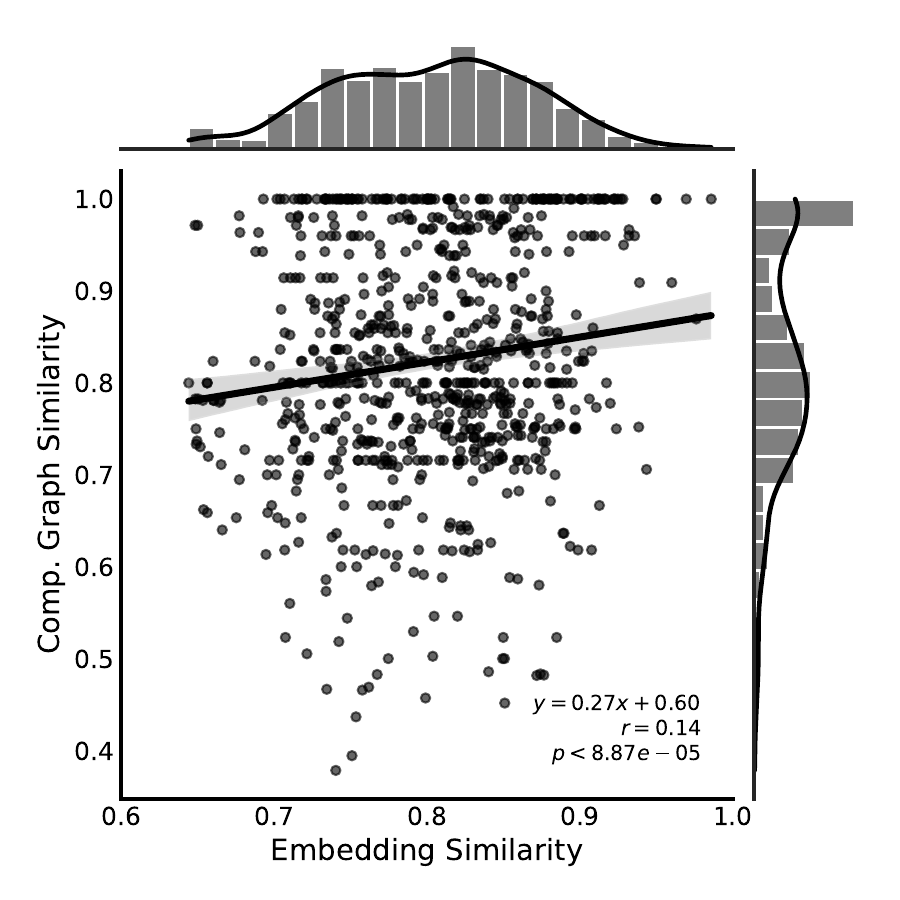}
    \end{minipage}
    \caption{The scatter plot of our trained retriever (left) and BGE (right) on 100 random samples from Math23k. There is a stronger positive correlation between computational graph similarity and embedding similarity for data with trained retriever than the BGE model.}
    \label{fig:correlation}
\end{figure*}

Next, we present a case study on the retrieved data from Calc-ape210k using both our trained model and the BGE model. As shown in Figure \ref{fig:case_study}, for the query question, \textit{“The ‘Scientist’ series is 2.5 yuan/book, and the ‘Inventor’ series is 4 yuan/book. It costs a total of 22 yuan to buy two sets of books. There are 4 ‘Scientists’, how many books are there in the ‘Inventor’ series?”}, our trained retriever successfully retrieves examples with similar computational graphs, even though the semantics of these examples are quite different. In contrast, the original BGE model relies primarily on semantic similarity for retrieval. As illustrated in the figure, while all the retrieved questions in the BGE model relate to “books”, their computational graphs are entirely different from the query’s graph. Additionally, we include a scatter plot on the Math23k dataset, where we analyze the correlation between computational graph similarity and embedding similarity for the top-8 retrieved data points from 100 random samples, as depicted in Figure \ref{fig:correlation}. The results show that the Pearson correlation coefficient for our trained model is significantly higher than that for the BGE model, indicating that our approach is more effective in retrieving examples with similar computational graphs based on question embeddings.

\subsubsection{Performance with Different Amount of Training Data}

\begin{figure}[htbp]
    \centering
    \includegraphics[width=0.48\textwidth]{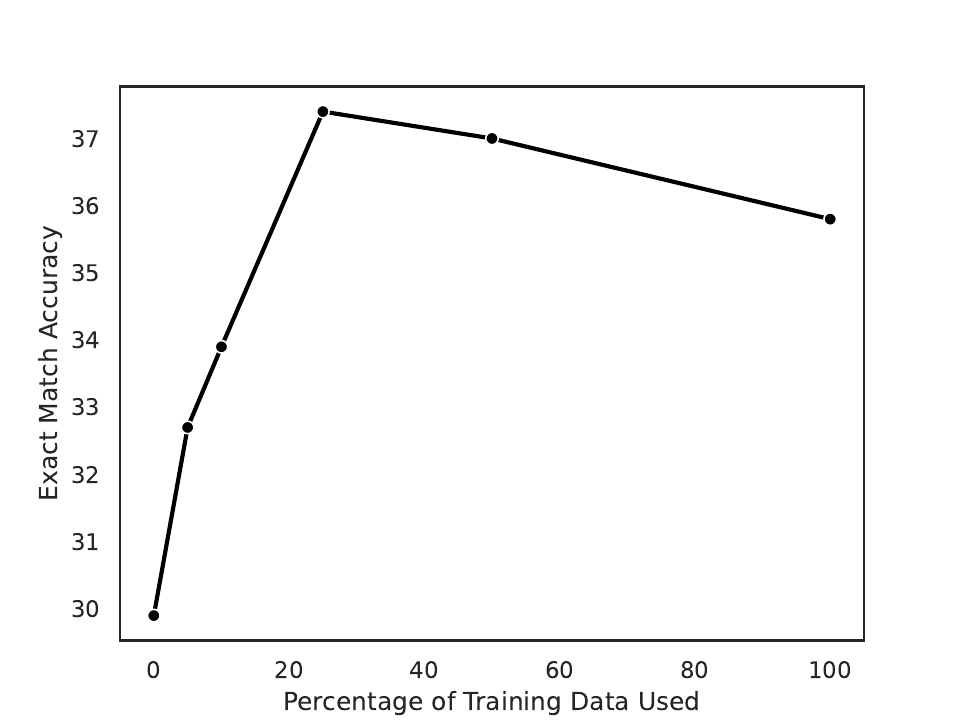} 
    \caption{The relationship between model performance on aqua\_rat dataset and the amount of data used. The performance stops to boost when using more than 25\% of training data. }
    \label{fig:percent}
\end{figure}

We also investigate how much training data is required to achieve optimal performance, given that manually labeling computational graphs can be expensive. To this end, we use 5\%, 10\%, 25\%, 50\%, and 100\% of the training data and examine the downstream performance on the aqua\_rat dataset. Figure \ref{fig:percent} illustrates the trend of the exact match score as a function of the amount of training data used. We observe that performance stops to boost when more than 25\% of the training data (approximately 5,000 samples) are used. Even with only 10\% of the training data (approximately 2,000 samples), our approach achieves a 4 percent accuracy gain over the BGE baseline. This indicates that our method can be effectively applied in scenarios with limited human-annotated data.

\subsubsection{Effect of Corpus Choice}
Finally, we investigate whether the choice of retrieval corpus affects performance. Specifically, we explore the case where data with the same distribution as the query are not available to serve as the corpus, a scenario that is common in real-world applications. In this experiment, we use the SuperCLUE-Math6 dataset \cite{xu2024supercluemath6gradedmultistepmath}, where only the test set is available, and select the training set from ape210k dataset as the retrieval corpus. The results, shown in Table \ref{table:SuperCLUE}, demonstrate that our approach remains effective even when the corpus and the query data do not share the same distribution. This suggests that, despite the different data distributions, our trained retriever can still find problems with similar computational graphs in the large ape210k corpus. This capability indicates that our method can be applied in a broad and flexible manner, making it suitable for various real-world scenarios.

\begin{table}[htbp]
\centering
\renewcommand{\arraystretch}{1.2} % 设置行距为1.2倍
\resizebox{\linewidth}{!}{
\begin{tabular}{lccc}
\hline
& \textbf{Ours} & \textbf{BGE} & \textbf{Random} \\
\hline
\textbf{SuperCLUE-Math6} & \textbf{27.2} & 20.6 & 18.6 \\
\hline
\end{tabular}
}
\caption{Results with Qwen2.5 0.5B model on SuperCLUE-Math6 test set. Here we use the training set of ape210k as the retrieval corpus, as the training set of SuperCLUE-Math6 is not availale.}
\label{table:SuperCLUE}
\end{table}
\section{Computational Graph-Free Training Data Acquisition}

\begin{table*}[htbp]
\centering
\renewcommand{\arraystretch}{1.2} 
\begin{tabular}{lccccc}
\hline
\textbf{} & \textbf{gsm8k} & \textbf{math\_qa} & \textbf{Calc-ape210k} & \textbf{aqua\_rat} & \textbf{Avg.} \\
\hline
\textbf{BGE}  & 38.7 & 45.9 & 20.4 & 29.9 & 33.7 \\
\textbf{Ours\textsubscript{w/ labeled data}}  & \textbf{40.7} & \textbf{47.3} & \textbf{31.3} & \textbf{37.4} & \textbf{39.2} \\
\hline
\textbf{Ours\textsubscript{w/ distillation data}} & \underline{39.4} & \underline{46.4} & \underline{27.5} & \underline{35.0} & \underline{37.1} \\
\hline
\end{tabular}
\caption{Results with training data distilled from GPT-4o with LLaMA-3.2-1B-Instruct generator. Retriever trained with distilled data outperforms the BGE baseline while underperforms the model trained with labeled data on all tasks.}
\label{table:distillation}
\end{table*}
Although our method is effective when minimal human-annotated data is available, we aim to eliminate the reliance on expensive human labor. Notably, in our training pipeline, we only need data pairs that contain either the same or different computational graphs, rather than requiring the computational graphs themselves. This allows us to avoid the explicit need for computational graph annotations. Instead, we can leverage large language models (LLMs), such as Claude-3.5 or GPT-4 \cite{openai2024gpt4technicalreport}, to generate training data. 

\begin{figure*}[htbp]
    \centering
    \includegraphics[width=0.9\textwidth]{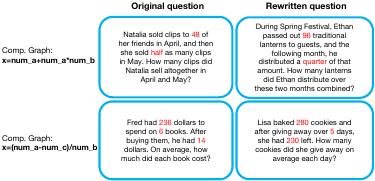} 
    \caption{Some cases of the original and rewritten questions. The entity names, value of numbers and semantics are different after rewritting, while the computational graphs remain the same.}
    \label{fig:rewritten}
\end{figure*}

To do this, we prompt the LLM to rewrite the questions so that all details, such as numerical values and entity names, differ from the original question, while maintaining the same computational graph. We use the following prompt of this rewritting: \textit{“Generate a problem with the same computation graph as the input math problem, ensuring that the semantics, numerical values, and sentence structure are as different as possible. Output only one rewritten example, without any additional information.”} We randomly select 5,000 samples from the training set of gsm8k and use this approach to generate 5,000 positive pairs to train the retriever. The downstream results, shown in Table \ref{table:distillation}, indicate that while the retriever trained with distilled data performs slightly below that trained with labeled data, it consistently outperforms the BGE baseline, demonstrating the effectiveness of the distilled data. Examples of this rewriting process are presented in Figure \ref{fig:rewritten}. Empirically, we observe that the sentence structure before and after rewriting is more similar than in the labeled data pairs, which the retriever may rely on to capture similarity between positive pairs during training, rather than focusing on the true computational graphs. We anticipate that more advanced methods will be developed in future work to construct high-quality training data without human labor.

\section{Related Work}

\paragraph{Few-shot Prompting for MWP Solving.} 
Large Language Models have shown promising results in tackling math word problems \cite{toshniwal24openmathinstruct, yang2024qwen2, DBLP:conf/iclr/YuJSYLZKLWL24, mirzadeh2024gsmsymbolicunderstandinglimitationsmathematical, wei2022chain}. To enhance model performance on math word problems, few-shot prompting has become a widely adopted approach \cite{wei2022chain, jiang2023mistral, melz2023enhancing, henkel2024retrieval}. The choice of examples used in few-shot prompting is critical to its success. Existing methods for example selection generally fall into two categories: semantic similarity-based retrieval \cite{huang2023boosting, melz2023enhancing, henkel2024retrieval} and random selection \cite{wei2022chain, jiang2023mistral, dubey2024llama3herdmodels}. By contrast, our approach leverages a computational graph-based retrieval strategy. Rather than relying solely on superficial linguistic features, our method retrieves examples that match the mathematical structure of the target problem. This structurally informed selection enables LLMs to draw from examples that better align with the mathematical reasoning required, enhancing the effectiveness of few-shot prompting for MWP solving.

\paragraph{Retrieval-Augmented Generation.} Retrieval-Augmented Generation (RAG) has recently gained attention to improve the quality of LLM outputs by integrating relevant external information during generation \cite{lewis2020retrieval, gao2023retrieval, fan2024survey}. In the context of math word problems, RAG has been applied to enhance the performance. Specifically, \citet{henkel2024retrieval} propose a RAG system by retrieving content from an open-source math textbook. Similarly, \citet{dixit2024sbi} introduced a schema-based RAG framework for math word problems, using structured schemas to guide LLMs in selecting appropriate mathematical operations, ultimately enhancing reasoning clarity and problem-solving structure. Our framework can also be viewed as a RAG system, where the corpus consists of structurally relevant MWP examples.

\paragraph{Contrastive Learning.} Contrastive learning has been adopted widely in various domains \cite{he2020momentum, chen2020simple, linpisces, DBLP:conf/iclr/LinXXW24, yu2022graph, DBLP:conf/acl/IterGLJ20, xu2022pisces}. Initially popularized in computer vision through methods like SimCLR \cite{chen2020simple} and MoCo \cite{he2020momentum}, contrastive learning has since been extended to other applications, such as text embedding tasks in NLP domains \cite{wang2023improving, behnamghader2024llmvec, wang2022text, lin2024unleashing}. The key idea behind contrastive learning is to pull together the embeddings of positive pairs while pushing apart those of negative pairs. In our work, we apply contrastive learning to train a retriever model designed for math word problems. Positive pairs in our approach consist of questions that share the same computational graph, capturing similar mathematical structures and reasoning patterns. Negative pairs are selected from other in-batch examples with different computational graphs. This training approach enables our retriever to identify structurally relevant examples during inference, thereby enhancing few-shot prompting and ultimately improving the performance of LLMs on math word problem-solving tasks.

\paragraph{Reasoning Ability in LLMs.} Large language models (LLMs) have often been criticized for lacking “system 2” thinking ability \cite{yu2024distilling21}, which limits their performance on complex reasoning tasks. Many prior studies have raised concerns about the “genuine” reasoning capabilities of current LLMs \cite{hazra2024largelanguagemodelsreason, wei2022emergentabilitieslargelanguage}, noting that LLMs struggle to distinguish between causality and correlation \cite{ashwani2024causeeffectlargelanguage} and are not strong abstract reasoners \cite{gendron2024largelanguagemodelsstrong}. These findings suggest that, despite their extensive pretraining on large-scale corpora, current LLMs are essentially pattern matchers \cite{mirzadeh2024gsmsymbolicunderstandinglimitationsmathematical}. While reasoning ability can be partially elicited through prompt engineering techniques like Chain-of-Thought \cite{wei2022chain}, this paper explores an alternative approach—providing the LLM with pre-existing reasoning paths rather than relying on the model to generate them independently. This approach may offer a more reliable and stable solution, as self-generated reasoning can often be fragile and prone to errors \cite{mirzadeh2024gsmsymbolicunderstandinglimitationsmathematical, li-etal-2024-deceptive}.

\section{Conclusion}

In this work, we have explored computational graph-based retrieval for solving math word problems, drawing inspiration from the analogy of reasoning paths between similarly structured problems. Our experiments on both English and Chinese math datasets demonstrate the effectiveness of our approach across models of different scales, with performance gains being more pronounced for smaller models. Additionally, by leveraging LLMs, we can automatically construct training data without relying on human labor. We hope this paper inspires future research on tackling a variety of reasoning tasks, extending beyond math word problems.

\section*{Acknowledgment} We thank Chi Han, Bangzheng Li and Daniel Elliott for the helpful discussions.

\bibliography{custom}

\appendix

\end{document}